\documentclass{article}

\usepackage[preprint, nonatbib]{neurips_2022}

\usepackage[utf8]{inputenc} 
\usepackage[T1]{fontenc}    
\usepackage{hyperref}       
\usepackage{url}            
\usepackage{booktabs}       
\usepackage{amsfonts}       
\usepackage{nicefrac}       
\usepackage{microtype}      
\usepackage{xcolor}         
\usepackage{minitoc}
\usepackage{amsmath}
\usepackage[ruled,linesnumbered]{algorithm2e}
\usepackage{graphicx}
\usepackage{multirow}
\usepackage{wrapfig}
\usepackage[at]{easylist}

\title{MRL: Learning to Mix with Attention and Convolutions}

\author{%
  Shlok Mohta\thanks{Corresponding Author.}
   , Hisahiro Suganuma, Yoshiki Tanaka \\
  Sony Group Corporation \\
  \texttt{\{shlok.mohta, hisahiro.suganuma, yoshiki.tanaka\}@sony.com} \\
}

\begin{document}

\maketitle

\begin{abstract}
In this paper, we present a new neural architectural block for the vision domain, named \textit{Mixing Regionally and Locally} (MRL), developed with the aim of effectively and efficiently \textit{mixing} the provided input features. We bifurcate the input feature mixing task as mixing at a regional and local scale. To achieve an efficient mix, we exploit the domain-wide receptive field provided by self-attention for regional-scale mixing and convolutional kernels restricted to local scale for local-scale mixing. More specifically, our proposed method mixes regional features associated with local features within a defined region, followed by a local-scale features mix augmented by regionally features. Experiments show that this hybridization of self-attention and convolution brings improved capacity, generalization (right inductive bias), and efficiency. Under similar network settings, MRL outperforms or is at par with its counterparts in classification, object detection, and segmentation tasks. We also show that our MRL-based network architecture achieves state-of-the-art performance for H\&E histology datasets. We achieved DICE of 0.843, 0.855, and 0.892 for Kumar, CoNSep, and CPM-17 datasets, respectively, while highlighting the versatility offered by the MRL framework by incorporating layers like group convolutions to improve dataset-specific generalization. 
\end{abstract}

\section{Introduction}

    Over the last decade, significant advancements in the neural network model architecture for computer vision through the use of convolutions \cite{AlexNet} and, more recently, through the use of self-attention based models like Transformers \cite{ViT, DBLP:journals/corr/Deit}, popularized by the application in natural language processing domain \cite{NIPS2017_3f5ee243}, have helped realize unprecedented success in the domain. Within these advancements, recurring themes of exploiting datasets' inductive bias \cite{DBLP:journals/corr/CohenW16, DSF-CNN}, feature mixing \cite{DBLP:journals/corr/MLP-Mixer, DBLP:journals/corr/Swin, DBLP:journals/corr/CSwin}, large datasets \cite{DBLP:journals/corr/JFT}, and corresponding ever-larger model parameters \cite{DBLP:journals/corr/MLP-Mixer, DBLP:journals/corr/CoAtNet, ViT, DBLP:journals/corr/Deit} have been driving innovation. \\
    \indent While models based on self-attention \cite{ViT} or MLP \cite{DBLP:journals/corr/MLP-Mixer}, offer parameter scaling to up to billions of parameters, higher model capacities, and are best suited to exploit a large dataset, they often come at the cost of sub-par performance in the low data regime compared to ConvNets counterparts given the same data and computational resources. Such behavior indicates a lack of desirable inductive biases in such models, which are learned parametrically from raw data. ConvNets, on the other hand, with their desirable inductive biases, are efficient on datasets of various sizes, offer a better generalization, and faster convergence, but lack scaling, being explored in recent works \cite{DBLP:journals/corr/ConvNext}. As noted in \cite{DBLP:journals/corr/CoAtNet}, various recent works explore architecture design, designed to exploit the best of both high modeling capacities of self-attention type modules and inductive biases of ConvNets through local receptive fields for computationally efficient attention \cite{DBLP:journals/corr/Swin, DBLP:journals/corr/CSwin, chen2022regionvit, DBLP:journals/corr/halonet}, or augmenting networks with explicit or implicit convolutional layers \cite{DBLP:journals/corr/CvT}, however, such approaches lack a systematic understanding of how such design features interact with each other for an efficient network design. While \cite{DBLP:journals/corr/CoAtNet} offers a systematic approach for hybridizing convolutions and attention from the aspect of generalization and model capacity, more first-principles analysis in terms of feature mixing, computational cost, robustness to transferring in low data regimes is required to create efficient layers and network architectures. \\
    \indent In this work, we fundamentally explore the idea of efficiently `mixing' input features by a neural network block. To this end, we take inspiration from \cite{DBLP:journals/corr/MLP-Mixer, DBLP:journals/corr/FNet}, suggesting feature mixing is sufficient for good performance and assimilating ideas from established techniques in the vision domain for achieving an efficient mixing of contextual information across the input domain \cite{chen2022regionvit}. Our experiments show that coarsely mixing features using self-attention on a global scale can supplement fine scales feature mixing using convolutions. Such an arrangement of layers provides essential global scale contextual information, with higher model capacity in a computationally efficient manner while allowing exploitation of the dataset's inherent symmetries, which conforms to better generalization and faster convergence speed attributed to strong priors of inductive bias. In this paper, we investigate the following two insights: Firstly, we observe that composing a block with self-attention and convolutions can be used as a drop-in replacement to the Multi-Head Attention block employed as part of various vision Transformer architectures, allowing for computationally efficient network architecture with no-to-minimal accuracy degradation; Secondly, based on the inherent symmetries presents in the datasets, the block augmented with layers utilizing such prior knowledge can facilitate better generalization, network convergence, and robustness to low data regimes. Based on these insights, we propose a simple yet effective novel architectural unit, termed as "Mixing Regionally and Locally" (MRL) block for building specialized hybrid vision neural architectures, with self-attention and convolutions at the core. \\
    \indent We set our priority to evaluate the proposed neural network block's ability to perform as a general-purpose architectural unit rather than put forward a new neural network architecture design based around it. To that end, our proposed MRL block is computationally efficient and achieves accuracy similar to the variants of Transformer architectures for which we replace the Multi-Head Attention block, responsible for feature mixing,  with the MRL block, highlighting the scalability offered by self-attention while at the same time enjoying the generalization and faster convergence properties offered by convolutions. Further to demonstrate that MRLs perform favorably on downstream tasks of classification, detection, and segmentation, requiring the ability for dense prediction, we compare MRL against current state-of-the-art methods under similar network constraints on COCO datasets. Also, for datasets with relatively small data sizes and strong symmetry priors, we evaluate MRL on multiple H\&E histology image datasets. We show that MRL-based backbone architectures achieve state-of-the-art performance on these datasets, a domain dominated by ConvNets of relatively small parameter sizes.

\section{Method}

\begin{figure}[ht]
    \centering
    \includegraphics[width=10cm]{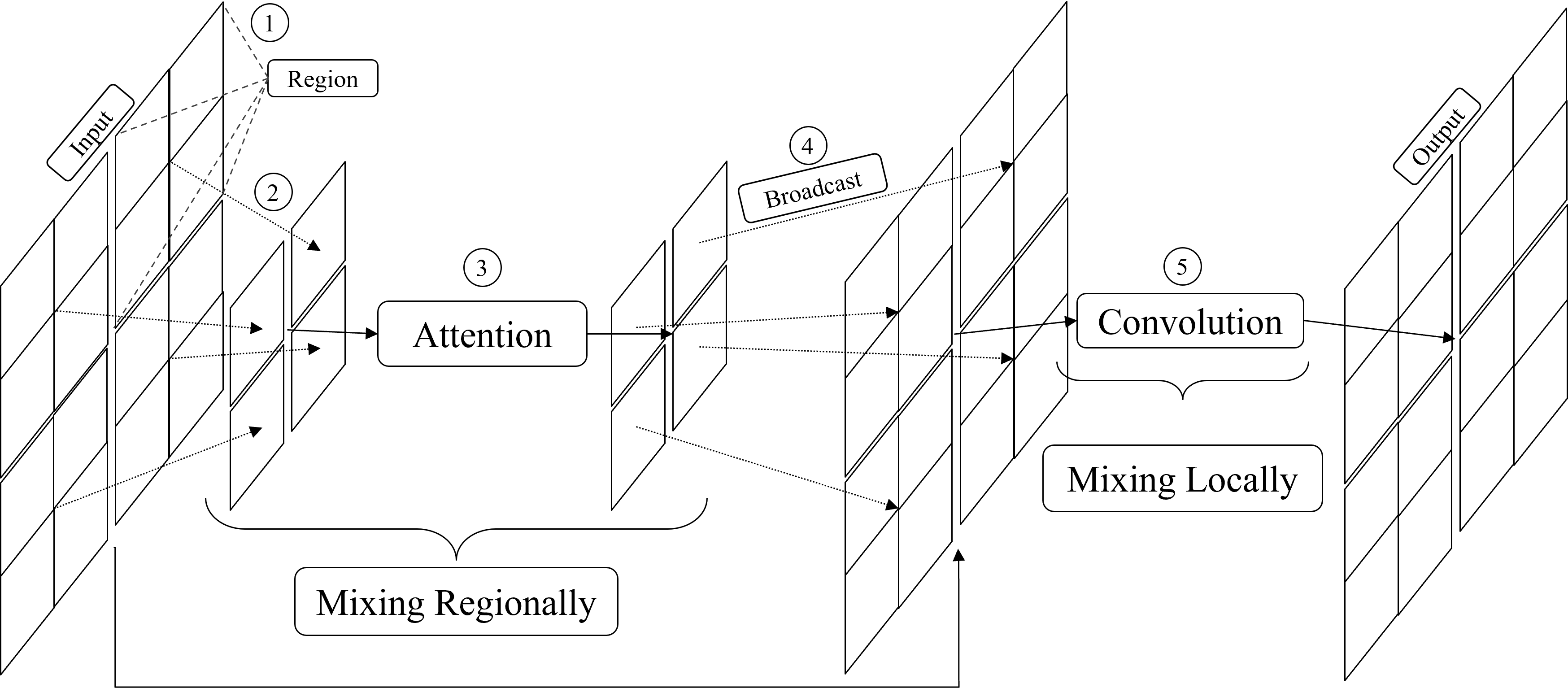}
    \caption{Schematic diagram representing Mixing Regionally and Locally approach. Each rectangle corresponds to an individual feature vector. Numbers in the diagram correspond to the respective steps described in Section \ref{subsec:MixRegion} and Section \ref{subsec:MixLocal}, as part of the MRL block.}
    \label{fig:MRL-Mech}
\end{figure}

In this section, we focus on how to optimally `mix' the provided input feature to a neural architecture unit, which we pose as the following questions to further our analysis of image-based inputs: 
\begin{enumerate}
    \item How to efficiently and effectively capture global scale information for a given input feature map?
    \item How to exploit datasets' inherent symmetries for achieving better generalization, augmented with access to global scale contextual information? 
\end{enumerate}
The above decomposition of the `mixing' task guides our analysis by doing a two-part process described in the Section \ref{subsec:MixRegion} and Section \ref{subsec:MixLocal}, respectively. Additionally, in \ref{subsec:implement}, we detail some practical considerations we experiment with while exploring MRL. As we further elucidate our design, our rationale for adopting the above decomposition of the task will become clear. Figure \ref{fig:MRL-Mech} presents a schematic diagram of our proposed method.

\subsection{Mixing Regionally}\label{subsec:MixRegion}

Considering an image as a signal consisting of multiple frequencies, to efficiently capture global scale information, a large receptive field would be crucial to capture essential contextual information embedded in the frequency components spanning across the width of the input signal. However, receptive fields spanning the entire spatial domain might be redundant considering a trade-off between model performance and associated computational cost. \\
\indent Focusing on optimizing the computational aspect of capturing relevant global scale structure, optimally processing the low-frequency component of the input signal by an efficient sampling of the input signal followed by processing with a receptive field spanning the entire sampled spatial domain would be crucial. Such processing on the input signal would minimize computational redundancy while enriching the network with high-level concepts for enhanced performance. \\
For any input of size $n \times n$ (channel dimension omitted for clarity), we devise the following step for efficiently capturing global scale information. \label{steps:mixr}
\begin{enumerate}
    \item Group input features into \textit{regions} \\
    This step is akin to preparing the input signal for sampling the frequency components. Based on a sampling window size $r < n$, a design hyperparameter, the input is partitioned into patches of $r \times r$. For the current study, the sampling windows that divide the input features are strictly non-overlapping. However, overlapping sampling windows can also be utilized but not studied as part of the current study. Also, as proposed in \cite{DBLP:journals/corr/Swin}, a \textit{shifted window} scheme can also be employed to allow for cross-window connection. We refer to the input partitioned by the sampling window as \textit{region}.
    \item Sample down each \textit{region}. \\
    This step is akin to sampling the region. Distilling the \textit{region} to a singular feature vector, using a learned convolutional kernel with a size equivalent to sampling window size, $r$, is performed as part of the sampling. This step is crucial as this helps reduce the input space to a new, $\frac{n}{r} \times \frac{n}{r}$, \textit{sampled input}. We use a convolution layer for this step, with kernel size and stride of $r$, due to highly optimized compute kernels, but down-sampling techniques like max/min pooling, and bilinear downsampling, may work in practice.
    \item \label{step:3} Process the sampled input with a large receptive field. \\
    Finally, as the initial input signal has been down-sampled, processing the input with a large receptive field is computationally efficient. For this purpose, self-attention with its receptive field spanning the entire spatial location fits the bill, as it allows for capturing input-dependent complex relational interactions, a desirable property when processing high-level concepts. 
\end{enumerate}

\subsection{Mixing Locally} \label{subsec:MixLocal}
Image contains strong local structures, like, spatially neighboring pixels are usually highly correlated, object symmetries, exploiting which is crucial for better generalization, faster convergence, and increased robustness for a given model. For this purpose, convolutions play the role perfectly as convolutional layers with their local receptive fields, shared weights, and spatial subsampling \cite{LeCun1999} are well suited for infusing information regarding texture and local structure \cite{DBLP:journals/corr/texture-bias}. Moreover, enhancing convolutional layers through group convolutions \cite{DBLP:journals/corr/CohenW16, DSF-CNN} to exploit symmetries other than translational symmetries, such as rotational symmetries, mirror symmetries, embedded in the local structure of the input can allow for networks that are \textit{less data hungry} with minimal parameter overhead. \\
\indent While convolutional layers perform well at a local scale, as noted previously, large receptive fields are essential for capturing more contextual information for higher model capacity. To that end, by combining the feature map generated as part of the \ref{subsec:MixRegion} to the feature map to input feature map,  $n \times n$ input, for extracting local features, we can bring in the advantages offered by both. \\
Continuing from the steps devised in Section \ref{subsec:MixRegion}, we propose the following steps for \textit{Mixing Locally} augmented with features containing global scale information, as a part of \textit{Mixing Regionally and Locally} block:
\begin{enumerate}
\setcounter{enumi}{3}
    \item Upscale-Sum feature map containing global scale information.\label{step:4} \\
    The feature map calculated as a part of the Mixing Regionally has a scale of $\frac{n}{r} \times \frac{n}{r}$, compared to the input $n \times n$. To augment the block input with global scale information, we \textit{broadcast} and \textit{sum} the individual feature vector obtained by Mixing Regionally to the features in their respective region of the block input $n \times n$. This step is akin to doing a \textit{nearest} neighbor upscaling of Mixing Regional feature map by a factor of $r$ and adding the result to the input to the block. 
    \item Process the augmented input feature with local receptive fields. \\
    Using convolutional layers with kernel sizes restricted to local scales, we process the input feature set to the block to extract local scale structures by taking advantage of the built-in inductive biases of convolutional layers to produce the block output. Depending on the inherent structural symmetries present in the datasets, utilizing group convolutions can allow the model to achieve better generalization and faster convergence.  
\end{enumerate}


\section{Related Works}
\paragraph{Mixing Features} 
Our fundamental analysis is aimed to gain a better understanding of how we can efficiently `mix' features. As presented by \cite{DBLP:journals/corr/MLP-Mixer}, they show that the `mixing' of input features, which they achieved by applying MLPs along and across input feature tokens, is sufficient for attaining high accuracies for computer vision tasks. Similarly, from the natural language processing domain, \cite{DBLP:journals/corr/FNet} shows that Transformer encoder architectures can be sped up, with limited accuracy costs, by replacing the self-attention sub-layers with simple linear transformations that "mix" input tokens. Our insights from these papers helped define the hypothesis that, at its base, models based on convolutions \cite{AlexNet, DBLP:journals/corr/ResNet, DBLP:journals/corr/efficient-net, DSF-CNN} and self-attention (vision-transformer models) \cite{ViT, DBLP:journals/corr/Deit, DBLP:journals/corr/pvt, DBLP:journals/corr/cait} are mixing features. While at the same time, they augment their output representation with their specific functional advantages, for example, input-independent invariance properties of convolutions and parameter independent scaling of receptive field of self-attention. In contrast convolutional, self-attention or hybrid approaches, our proposed neural network architecture unit stands to achieve the `mixing' of the input features as the unit's intended behavior, not as the consequences of its constituents. 
\paragraph{Using Regions}
A common recurring theme in various literature \cite{ViT, DBLP:journals/corr/halonet, DBLP:journals/corr/Twins, DBLP:journals/corr/standalone-atten, DBLP:journals/corr/Swin, DBLP:journals/corr/CSwin, chen2022regionvit} is to divide the input feature set into regions/patches, specifically while using self-attention, to improve the model's speed and memory usage. The standard vision transformer proposed in \cite{ViT} uses patches to control the computation-accuracy trade-off of using self-attention. \cite{DBLP:journals/corr/standalone-atten} replaces standard convolutional kernels with self-attention, restricting the receptive field of self-attention while allowing for increased computational performance. \cite{DBLP:journals/corr/Swin} purpose the approach of \textit{shifted windows} which brings in greater computational efficiency when utilizing self-attention by limiting the computation to non-overlapping local windows while allowing for a cross-window connection. \cite{DBLP:journals/corr/CSwin} purpose a similar window-based self-attention approach, wherein, they propose computing self-attention in the horizontal and vertical stripes in parallel, forming a \textit{cross-shaped} window. \cite{DBLP:journals/corr/Twins} performs a sub-sampling of the input feature set for computing self-attention, interleaving sub-sampled feature set self-attention block and full-feature set self-attention to improve network efficiency. 
While \cite{chen2022regionvit} adopts a strategy of utilizing self-attention to learn both regional and local scale features while augmenting the local scale self-attention through their respective processed regional tokens.
\paragraph{Convolutions \& Attention} Various previous works have explored combining convolutions and attention for designing neural network models to reap the benefits provided by both. \cite{2020Lite} presents the use of self-attention and convolutions for allowing the network to gain different perspective of global and local information while reducing the computational cost for NLP tasks. \cite{muse} presents a study where convolutions are used to augment self-attention with stronger local information for long sequences. For vision based tasks, one line of exploration is augmenting existing ConvNet backbones with self-attention modules \cite{DBLP:journals/corr/botnet, DBLP:journals/corr/attenAugConv} or replacing convolutional layers with self-attention \cite{DBLP:journals/corr/halonet, DBLP:journals/corr/standalone-atten}. Another line of research explores augmenting Transformer based backbones with convolutions \cite{DBLP:journals/corr/CvT, DBLP:journals/corr/T2T}, trying to infuse desirable qualities of convolutions into successful self-attention-based architectures \cite{ViT}. While these approaches focus on augmenting networks with convolutions and attention, in an ad-hoc manner, \cite{DBLP:journals/corr/CoAtNet} explore the combining self-attention and convolutional block within one basic computational block.

\BlankLine
While our work belongs to this category and shares certain design aspects with \cite{DBLP:journals/corr/Twins, chen2022regionvit}, and comes closest to \cite{DBLP:journals/corr/CoAtNet, muse} in terms of combining convolutions and self-attention as a part a one computational block, our proposed neural architectural unit is, by design, an elementary way of mixing global scale and local scale information. Like \cite{DBLP:journals/corr/CoAtNet}, MRL inherits the generalization properties inherited by ConvNets, and scalability offered by self-attention while at the same the has a lower computational footprint for self-attention, making it suitable for high-resolution inputs. We have provided a detailed list of differentiating factors between our proposed MRL and other techniques utilizing Convolutions and Attention in Appendix \ref{Appen:Related_work}.

\section{Experiments}

In this section, we stand to evaluate the MRL block's ability to serve as an efficient neural network architectural block and establish its utility for practical tasks. For our evaluation, we replace the Multi-Head Self-Attention (MHSA or simply SA) module in existing Transformer architectures with an MRL block and evaluate if the network maintains similar network accuracy. Replacing MHSA with MRL helps verify the hypothesis that self-attention is one way of `mixing' features, hence can be replaced by MRL, and showing that MRL can preserve the high modeling capacity offered by self-attention based neural network models. To evaluate MRL, we decompose our evaluation criteria into two categories, described as follows:

\begin{enumerate}
    \item Section \ref{subsec:imagenet}: Ability to serve the role of a general-purpose neural network architecture unit.
    \item Section \ref{subsec:COCO}, \& \ref{subsec:MBG}: Ability to perform favorably on downstream tasks.
\end{enumerate}
For any neural network, the ability to perform well on various downstream tasks is essential to validate its utility. To that end, we focus on tasks that require dense predictions while posing practical challenges in terms of dataset size and exploiting dataset objects' symmetries. We also evaluate how MRL's design choices impact the network's generalization and convergence. We also discuss potential limitations of MRL on such tasks.

\subsection{ImageNet Classification}\label{subsec:imagenet}

\subsubsection{Setup}
\paragraph{Base Network} We replace the MHSA module in CvT\footnote{https://github.com/microsoft/CvT}, the network architecture proposed by \cite{DBLP:journals/corr/CvT}, with the MRL module. We adopt CvT as our base network for the following reasons: 1) A fully self-attention based network architecture, effective in highlighting the effects of replacing the MHSA module with MRL; 2) Multiple model variants, varying by parameters and FLOPs, allowing for evaluating MRL's ability to scale with deeper networks; 3) Use of convolutional embeddings for a simplified adaption to various vision tasks, blending well with the input-outputs for the MRL module. The family of models used for this evaluation, replacing the MHSA with the MRL module, based on the CvT family of models, is described in Table \ref{table:cvtmodel} of Appendix \ref{Appen:IC}. Except for the MHSA module of CvT, the other network design parameters are the same. Also, from the MRL side, considering the network's intermediate window sizes, we use region window sizes with a multiple of 2 for ease of evaluation. 
\vspace{-4mm}
\paragraph{Training \& Finetuning} We train the models specified in Table \ref{table:cvtmodel} on the ImageNet dataset with 1.3M images
. For the individual network specified in Table \ref{table:cvtmodel}, we use training settings similar to \cite{DBLP:journals/corr/CvT}, not limited to the number of training epochs, optimizer, and weight decay, allowing us to compare our proposed MRL one-on-one against purely SA blocks. We adopt data augmentation and regularization methods similar to ViT \cite{ViT}. Unless stated otherwise, an input of $224 \times 224$ is utilized. 
Similarly, we employ the same fine-tuning strategies, along with fine-tuning hyperparameters, as in \cite{DBLP:journals/corr/CvT}, where-in we train the models on an input size of $224 \times 224$, from ImageNet-1K
, and fine-tune the models on an input of $384 \times 384$ of ImageNet dataset. 

\subsubsection{Results}

Table \ref{table:Imagentresults} presents the summarizes and compares the results of training CvT models, replacing the SA block with the MRL block. MRL model variants achieve Top-1 accuracies similar to or higher than their CvT model counterparts, with up to 50\% FLOPs , corresponding to replaced part of the network, and 12\% training throughput improvement for the pre-training phase where an input of $224 \times 224$ is utilized. MRL brings significant FLOP and training throughput improvements for larger inputs, as is the case for inputs of $384 \times 384$, for finetuning. MRL achieves finetuning accuracies similar to its CvT counterparts with up to 30\% improvement in network training throughput and 70\% FLOPs improvement in replaced part of the network. Moreover, the reduced memory consumption of the MRL block compared to SA allows for using a larger mini-batch size per GPU (executed on A100 GPU with 40GB memory), allowing for further throughput improvement. 

\begin{table}[htbp]
  \centering
  \caption{Model performance (Top-1 classification accuracy) on ImageNet. $1K$ denotes training on ImageNet-1K; {\fontfamily{cmss}\selectfont $1K_{384}$} denotes pre-training on ImageNet-1K followed by finetuning on ImageNet-1K with an input size of 384; {\fontfamily{cmss}\selectfont SA/MRL} denotes the Parameter and FLOPs corresponding to SA or MRL blocks of the network; {\fontfamily{cmss}\selectfont  images/s} correspond to \textbf{training throughput} on one DGX-A100 Node, with the subscript denoting the mini-batch size used per GPU (provided to highlight cases where a large-batch size could could not be processed for CvT models due to out-of-memory errors).} \label{table:Imagentresults}
    \resizebox{0.7\columnwidth}{!}{%
    \begin{tabular}{c|l|cc|cc|c|c}
    \toprule
    \multicolumn{1}{c}{} &       & \multicolumn{2}{c|}{\#Paras (M)} & \multicolumn{2}{c|}{FLOPs (G)} & \multicolumn{1}{c|}{Top-1} & \\
    \multicolumn{1}{c|}{Dataset} & \multicolumn{1}{c|}{Model} & SA/MRL & Full  & SA/MRL & Full  & \multicolumn{1}{c|}{Accuracy} & images/s \\
    \midrule
    \multirow{4}[4]{*}{$1K$} & \multicolumn{1}{c|}{CvT-13} & 6.40  & 19.98 & 1.42  & 4.50  & 81.6  & 496.4 \\
          & \multicolumn{1}{c|}{MRL-13} & 6.40  & 19.98 & 0.63  & 3.71  & \textbf{81.63} & \textbf{555.9} \\
\cmidrule{2-8}          & \multicolumn{1}{c|}{CvT-21} & 10.34 & 31.54 & 2.17  & 7.13  & 82.3  & 308.6 \\
          & \multicolumn{1}{c|}{MRL-21} & 10.35 & 31.55 & 1.03  & 5.98  & \textbf{82.3} & \textbf{346.3} \\
    \midrule
    \multirow{4}[3]{*}{$1K_{384}$} & \multicolumn{1}{c|}{CvT-13} & 6.40  & 19.98 & 7.10  & 16.30 & 83    & 132.2 \\
          & \multicolumn{1}{c|}{MRL-13} & 6.40  & 19.98 & 1.99 & 11.19 & \textbf{83.1} & \textbf{172.4} \\
\cmidrule{2-8}          & \multicolumn{1}{c|}{CvT-21} & 10.34 & 31.54 & 10.20 & 24.90 & 83.3  & $81.7_{32}$ \\
          & \multicolumn{1}{c|}{MRL-21} & 10.35 & 31.55 & 3.23  & 17.93 & \textbf{83.3} & \boldmath{}\textbf{$107.8_{64}$}\unboldmath{} \\
    \end{tabular}%
    }
\end{table}%
\paragraph{Experimentation Limit.} One limitation of the presented experiments is that we have not presented results strongly relating MRL-based networks \textit{modeling capacity}, by performing experiments with larger models (number of parameters) and larger datasets like ImageNet-21k \cite{Imagenet}. However, we have highlighted MRL-based networks' higher modeling capacity in Section \ref{subsec:MBG}, by showing MRL's ability to exploit extra datasets, leading to better performance than its competition. Also, we have not presented the study evaluating the effect of larger/smaller region size and of increasing the number of filters per channel for the depthwise convolutions in the MRL block. We leave that as a future work.

\subsection{COCO Object Detection}\label{subsec:COCO}
Next, we evaluate our MRL block on the COCO object detection task with the Mask R-CNN \cite{Mask-RCNN} and the Cascade Mask R-CNN \cite{Cascade_maskrcnn} framework. Specifically, since we do not propose our own network design, that is the number of network stages or the number if MRL blocks per stage, we use the \textbf{Swin-T and Swin-S \cite{DBLP:journals/corr/Swin}\footnote{https://github.com/SwinTransformer/Swin-Transformer-Object-Detection} as backbone and only replace the self-attention mechanism with MRL} and compare our results against the same. We employ the same pre-training and finetuning strategy as used in Swin Transformer \cite{DBLP:journals/corr/Swin} on the COCO training set. 

\begin{table}[htbp]
  \centering
  \caption{Object detection and instance segmentation performance on the COCO val2017 with Mask R-CNN framework. The {\fontfamily{cmss}\selectfont FLOPs} and {\fontfamily{cmss}\selectfont FPS} (Frames per second) are calculated for an input size of $800 \times 1200$, on an A100 GPU. {\fontfamily{cmss}\selectfont MRL} is the model corresponding to networks with Swin Transformer attention replaced with MRL blocks.} 
  \resizebox{\columnwidth}{!}{%
    \begin{tabular}{c|c|c|c|ccc|ccc|ccc|ccc}
    \toprule
          & \#Params & FLOPs & \multirow{2}[2]{*}{FPS} & \multicolumn{6}{c|}{Mask R-CNN 1$\times$ schedule}   & \multicolumn{6}{c}{Mask R-CNN 3$\times$ + MS schedule} \\
          & (M)   & (G)   &       & AP$^b$ & AP$^{b}_{50}$ & AP$^{b}_{75}$ & AP$^m$ & AP$^{m}_{50}$ & AP$^{m}_{75}$ & AP$^b$ & AP$^{b}_{50}$ & AP$^{b}_{75}$ & AP$^m$ & AP$^{m}_{50}$ & AP$^{m}_{75}$ \\
    \midrule
    Swin-T & 48    & 267   & 25.2  & 43.7  & 66.6  & 47.7  & 39.8  & 63.3  & 42.7  & 46.0  & 68.1  & 50.3  & 41.6  & 65.1  & 44.9 \\
    MRL-T & 50    & 251   & \textbf{25.4} & \textbf{45.2} & \textbf{68.2} & \textbf{49.4} & \textbf{41.1} & \textbf{65.0} & \textbf{44.0} & \textbf{46.5} & \textbf{68.6} & \textbf{51.2} & \textbf{42.0} & \textbf{65.4} & \textbf{45.2} \\
    \midrule
    Swin-S & 69    & 354   & 19.1  & 44.8     & 66.6     & 48.9     & 40.9     & 63.4     & 44.2     & 48.5  & 70.2  & 53.5  & 43.3  & 67.3  & \textbf{46.4} \\
    MRL-S & 72    & 322   & \textbf{19.2} & \textbf{45.7}     & \textbf{67.9}     & \textbf{50.1}     & \textbf{41.1}     & \textbf{64.6}     & \textbf{44.4}     & \textbf{48.5}     & \textbf{70.3}     & \textbf{53.5}     & \textbf{43.3}     & \textbf{67.3}     & 46.3 \\
    \end{tabular}%
    }
  \label{tab:COCO_maskrcnn}%
\end{table}%

\begin{table}[htbp]
  \centering
  \caption{Object detection and instance segmentation performance on the COCO val2017 with Cascade Mask R-CNN framework.}
  \resizebox{\columnwidth}{!}{%
    \begin{tabular}{l|c|c|c|ccc|ccc|ccc|ccc}
    \toprule
          & \#Params & FLOPs & \multirow{2}[2]{*}{FPS} & \multicolumn{6}{c|}{Cascade Mask R-CNN 1$\times$ schedule} & \multicolumn{6}{c}{Cascade Mask R-CNN 3$\times$ + MS schedule} \\
          & (M)   & (G)   &       & AP$^b$ & AP$^{b}_{50}$ & AP$^{b}_{75}$ & AP$^m$ & AP$^{m}_{50}$ & AP$^{m}_{75}$ & AP$^b$ & AP$^{b}_{50}$ & AP$^{b}_{75}$ & AP$^m$ & AP$^{m}_{50}$ & AP$^{m}_{75}$ \\
    \midrule
    Swin-T & 86    & 745   & 14.4  & 48.1  & 67.1  & 52.2  & 41.7  & 64.4  & 45.0  & 50.4  & 69.2  & 54.7  & 43.7  & \textbf{66.6}  & 47.3 \\
    MRL-T & 88    & 732   & \textbf{14.4} & \textbf{49.5} & \textbf{68.8} & \textbf{53.4} & \textbf{42.9} & \textbf{66.1} & \textbf{46.4} & \textbf{50.4} & \textbf{69.3} & \textbf{54.7} & \textbf{43.7} & 66.5 & \textbf{47.5} \\
    \midrule
    Swin-S & 107   & 833   & 11.9  & -     & -     & -     & -     & -     & -     & \textbf{51.9}  & \textbf{70.7}  & \textbf{56.3}  & \textbf{45.0}  & \textbf{68.2}  & \textbf{48.8} \\
    MRL-S & 110   & 800   & \textbf{12} & -     & -     & -     & -     & -     & -  & 51.6 & 70.4 & 56.1 & 44.9 & 68.0 & 48.6 \\
    \end{tabular}%
    }
  \label{tab:COCO_cascade}%
\end{table}%

\noindent Table \ref{tab:COCO_maskrcnn} reports the results of the Mask R-CNN framework with "$1\times$" (12 training epochs) and "3$\times$ + MS" (36 training epochs with multi-scale training). It shows that MRL variant of the network outperforms or performs at par with its Swin Transformer counterparts, with \textbf{+1.5} box AP, \textbf{+1.3} mask AP with 1$\times$ schedule for Tiny (\textit{T}) variant. This increase in accuracy in 1$\times$ schedule points towards better faster convergence and better generalization provided by convolutions, part of Mixing Locally in the MRL blocks. We also achieve similar accuracy statistics on the Small (\textit{S}) 1$\times$ schedule configuration. While for 3$\times$ schedule, the MRL variant performs at par with its Swin counterpart. \\
\noindent Table \ref{tab:COCO_cascade} presents the results of the Cascade Mask R-CNN framework. We observe a similar trend to the results of Mask R-CNN in Table \ref{tab:COCO_maskrcnn}, wherein the MRL variant surpasses or is at par with its Swin variants while maintaining equivalent throughput.

\noindent Additionally, we compare MRL block performance against existing variants of the self-attention mechanism used in the vision domain on the COCO dataset. We do this as MRL-based networks can be considered one of the variants of Vanilla Vision Transformer. Following the comparison setting in \cite{DBLP:journals/corr/Swin}, we use \textbf{Swin-T as the backbone and change the self-attention mechanism}. The result of the comparison using the Mask R-CNN framework, trained on 1$\times$ schedule is reported in Table \ref{tab:COCO_compare}. While our MRL network variant does not achieve the highest Top1 accuracy on the ImageNet-1K dataset used for pretraining the network, our MRL variant outperforms other self-attention variants on this task, highlighting better generalization offered by convolutions embedded in the MRL block. 

\begin{table}[htbp]
  \centering
  \caption{Comparison between different self-attention mechanism on COCO dataset, using Mask R-CNN framework. Results taken from \cite{DBLP:journals/corr/CSwin}.}
  \resizebox{0.5\columnwidth}{!}{%
    \begin{tabular}{l|c|cc}
    \toprule
          & \multicolumn{1}{c|}{ImageNet} & \multicolumn{2}{c}{COCO} \\
          & Top1 (\%) & AP$^b$ & AP$^m$ \\
    \midrule
    Shifted window \cite{DBLP:journals/corr/Swin} & 81.3  & 43.7  & 39.8 \\
    Spatially Sep \cite{DBLP:journals/corr/Twins} & 81.5  & 42.7  & 39.5 \\
    Sequential Axial \cite{axial} & 81.5  & 40.4  & 37.6 \\
    Criss-Cross \cite{ccnet} & 81.7  & 42.9  & 39.7 \\
    Cross-shaped window \cite{DBLP:journals/corr/CSwin} & \textbf{82.2} & 43.4  & 40.2 \\
   \textbf{MRL}  & 81.4  & \textbf{45.2} & \textbf{41.1} \\
    \end{tabular}%
    }
  \label{tab:COCO_compare}%
\end{table}%

\subsection{Histopathology: Instance Segmentation}\label{subsec:MBG}

\paragraph{Task \& Motivation} This current task pertains to performing nuclear level instance segmentation of Haematoxylin and Eosin (H\&E) stained histology slides.  Whole-Slide Images (WSIs) obtained by digitizing glass histology slides using optical scanning devices contains thousands of nuclei of various types. Assessing such WSIs can help predict clinical outcomes, for example, survivability \cite{10.1117/tumordiff} and diagnosing the grade and disease type \cite{Lu2018-as}. 
The following aspects of the current task motivated our analysis for the use of MRL based neural network architecture: 1) A dense prediction task, sufficient to highlight MRL's ability to perform as a neural network backbone architectural unit; 2) Relatively small dataset sizes for training/finetuning purposes, making model generalization a challengiing task; 3) Large neural network inputs and corresponding large intermediate neural network outputs, requiring large computational resources which can be alleviated by using MRL; 4) Strong symmetry bias in the dataset, exploiting which is necessary for better generalization and efficient models; 5) This task is dominated by ConvNet type architectures \cite{DBLP:journals/corr/hovernet, DSF-CNN, cdnet, Brp-net}, with limited progress with self-attention based architectures for histopathology \cite{DBLP:journals/corr/GasHis, shao2021transmil, transpath}.  

\subsubsection{Setup}

\paragraph{Base Network}
For our analysis, we use the network architecture, \textit{HoVerNet} \footnote{https://github.com/vqdang/hover\_net}, proposed in \cite{DBLP:journals/corr/hovernet}, as our base network architecture design for its state-of-the-art results on nuclear segmentation task. HoverNet, at its base, is a U-Net \cite{UNet} style neural network architecture with a base encoder network and decoder network. A detailed description of the HoVerNet architecture is presented in Appendix \ref{Appen:HoverNet}. We utilize MRL-based encoder and decoder architecture in HoVerNet. The architecture described in \cite{DBLP:journals/corr/hovernet} used a Preact-ResNet50 \cite{preactrestnet} as the encoder architecture and a series of up-sampling operations and densely connected units \cite{densecnn} for decoder networks, as shown in Figure \ref{fig:hovernet_arch} of the Appendix. We replace the encoder with MRL blocks-based architecture, with the same neural network units as in MRL networks described in Table \ref{table:cvtmodel}. Similarly, we replace the decoder networks with a series of up-sampling and MRL blocks. We refer to the this neural network architecture as \textit{MHVN}. Additionally, to exploit the inherent rotational symmetries present in these datasets \cite{DSF-CNN} we augment our MRL blocks by appending group convolutions \cite{DBLP:journals/corr/CohenW16} to the Mixing-Locally part, referring the resultant neural network architecture as \textit{GC-MHVN}. The network, along with the network parameters, are presented in Appendix.
\vspace{-4mm}
\paragraph{Dataset} As part of our analysis, we use the following datasets to perform instance segmentation evaluation: 
\textit{CoNSep} \cite{DBLP:journals/corr/hovernet}, \textit{Kumar} \cite{kumar}, and \textit{CPM-17} \cite{CPM17}. Additionally, we use \textit{Lizard} dataset, \cite{graham2021lizard}, as an extra \textit{pre-training} dataset, detail for which would be elucidated in the following sections. A full summary of these datasets can be found in Appendix \ref{appendix:datasets}. For dataset processing, we utilize similar data-processing pipelines as proposed in \cite{DBLP:journals/corr/hovernet} for all the dataset, that is, splitting the large WSIs into patches, which are then used as inputs to the network after corresponding data-augmentation. 
\vspace{-4mm}
\paragraph{Training \& Finetuning} Similar to \cite{DBLP:journals/corr/hovernet}, we pre-train the encoder part of the network with ImageNet-1K, utilizing all the same training and data-augmentation hyperparameters as described in \cite{DBLP:journals/corr/CvT}, which were also used for training the networks in Section \ref{subsec:imagenet}. However, to further exploit the model capacity offered by the self-attention module in the MRL block, we train both the encoder (pre-trained on ImageNet-1K) and decoder simultaneously on the Lizard dataset. For this step, we use the same training hyperparameters as used by \cite{DBLP:journals/corr/hovernet} for training the encoder and decoder simultaneously, however, we found that using warm-up for five epochs helped stabilize the training. Also, as the Lizard dataset is a culmination of other smaller datasets, including the CoNSep dataset, we remove all data corresponding to the CoNSep dataset while pre-processing the Lizard dataset for this phase.
We finetune the network pre-trained on Lizard(- CoNSep) dataset on CoNSep, Kumar and CPM17 dataset. We maintain the same training and test splits for all the datasets as used in \cite{DBLP:journals/corr/hovernet}. For this finetuning phase, we finetune the encoder and decoder parts of the networks simultaneously, using the same finetuning hyperparameters as utilized by \cite{DBLP:journals/corr/hovernet}, except we use a warm-up phase for five epochs for stabilizing the training. Also, to fully characterize and understand the performance of the proposed method, we use the evaluation metrics of \textit{DICE} score, Aggregated Jaccard Index (\textit{AJI}), \& Panoptic Quality (\textit{PQ}), as utilized by \cite{DBLP:journals/corr/hovernet}.

\subsubsection{Results}

\begin{table}[htbp]
  \centering
  \caption{Comparative experiments on Kumar \cite{kumar}, CoNSep \cite{DBLP:journals/corr/hovernet}, and CPM-17 \cite{CPM17} datasets. $Liz$ denotes network pre-trained on Lizard \cite{graham2021lizard} dataset.}\label{table:historesults}
  \resizebox{0.7\columnwidth}{!}{%
    \begin{tabular}{c|ccc|ccc|ccc}
    \toprule
    \multicolumn{1}{r}{} & \multicolumn{3}{c|}{Kumar} & \multicolumn{3}{c|}{CoNSep} & \multicolumn{3}{c}{CPM-17} \\
\cmidrule{2-10}    Method & DICE  & AJI   & PQ    & DICE  & AJI   & PQ    & DICE  & AJI   & PQ \\
    \midrule
    U-Net\cite{UNet} & 0.758 & 0.556 & 0.478 & 0.724 & 0.482 & 0.328 & 0.813 & 0.643 & 0.578 \\
    Mask-RCNN\cite{Mask-RCNN} & 0.760 & 0.546 & 0.509 & 0.740 & 0.474 & 0.460 & 0.850 & 0.684 & 0.674 \\
    DIST\cite{DIST}  & 0.789 & 0.559 & 0.443 & 0.804 & 0.502 & 0.398 & 0.826 & 0.616 & 0.504 \\
    Micro-Net\cite{micro-net} & 0.797 & 0.560 & 0.519 & 0.794 & 0.527 & 0.449 & 0.857 & 0.668 & 0.661 \\
    HoVer-Net\cite{DBLP:journals/corr/hovernet} & 0.826 & 0.618 & 0.597 & 0.853 & 0.571 & 0.547 & 0.869 & 0.705 & 0.697 \\
    \midrule
    DSF-CNN\cite{DSF-CNN} & 0.826 & -     & 0.597 & -     & -     & -     & -     & -     & - \\
    CDNet\cite{cdnet} & -     & -     & -     & 0.853 & 0.571 & -     & 0.880 & 0.733 & - \\
    BRP-Net\cite{Brp-net} & -     & 0.642 & -     & -     & -     & -     & 0.877 & 0.731 & - \\
    CIA-Net\cite{cia-net} & 0.818 & 0.620 & 0.577 & -     & -     & -     & -     & -     & - \\
    \midrule
    Pre-ResNet50$_{Liz}$ & 0.798 & 0.584 & 0.563 & 0.844 & 0.543 & 0.519 & 0.880 & 0.724 & 0.706 \\
    BiT101$_{Liz}$ & 0.826 & 0.623 & 0.604 & 0.848 & 0.570 & 0.554 & 0.882 & 0.726 & 0.715 \\
    MHVN$_{Liz}$ & 0.830 & 0.624 & 0.599 & 0.849 & 0.569 & 0.554 & 0.885 & 0.734 & 0.721 \\
    GC-MHVN$_{Liz}$ & \textbf{0.843} & \textbf{0.652} & \textbf{0.625} & \textbf{0.855} & \textbf{0.576} & \textbf{0.559} & \textbf{0.892} & \textbf{0.743} & \textbf{0.733} \\
    \end{tabular}%
    }
\end{table}%

\noindent Table \ref{table:historesults} compares our MRL-based models to other segmentation approaches used in computer vision \cite{Mask-RCNN}, medical imaging \cite{UNet}, and other approaches specifically tuned for nuclear segmentation \cite{DBLP:journals/corr/hovernet, Brp-net, cia-net, micro-net, DSF-CNN}. We present results of HoVerNet architecture pre-trained on Lizard dataset, \cite{DBLP:journals/corr/hovernet} with ImageNet-1K pretrained Pre-ResNet-50 \cite{preactrestnet} and BiT101 \cite{bitM} as encoders architectures. This allows us to compare and highlight the high \textit{model capacity} offered by self-attention in MRL blocks, for networks with equivalent parameter complexity. For final evaluation, we select the model corresponding to the epoch with the best DICE value on the validation set. \\
\indent The comparison table shows that the MRL-based model, \textit{GC-MHVN}, outperforms previous state-of-the-art approaches with a significant margin on all three datasets. In particular, we observe the following two characteristics of MRL, considering its performance on this downstream task. Firstly, MRL based HoVerNet architecture successfully exploits the extra training data available through the Lizard dataset, performing better than its ConvNet-HoVerNet counterparts with access to similar training data. It highlights MRL's ability to harness the higher model capacity offered by self-attention. Secondly, the higher performance of \textit{GC-MHVN} over \textit{MHVN} points towards MRL's ability to exploit the inductive biases offered by convolutions, in this case, rotational inductive bias by group convolutions, for better generalization. It highlights the dataset-dependent design versatility offered by the MRL unit framework. Both of the above characteristics indicate MRL framework's ability to combine self-attention and convolutions for better capacity and generalization. 

\paragraph{Limitation.} In our testing, we observe that MRL-based networks (both MHVN and GC-MHVN) during finetuning phase converge to high validation accuracies much faster, that is, within the first few epochs, compared to its ConvNet counterparts. This faster convergence is followed by a continuous improvement in the overall training loss, with a gradual drop in validation accuracies. This points toward over-fitting, commonly faced in self-attention-based models on limited datasets. Currently, we do not explore training settings to allow MRL-based models to continue learning relevant features, making post-epoch validation evaluation crucial depending on the dataset. 

\section{Conclusion}

In this paper, we have introduced a new neural network computational block named Mixing Regionally and Locally block (\textit{MRL}). The core design of the block is based on the idea of mixing features, which we decomposed as a task of mixing features at regional and local scales, to promote computational efficiency while maintaining the block's effectiveness. This mixing was achieved by exploiting self-attention and convolutions for mixing at the regional and local scales, respectively. The use of self-attention and convolutions provide models with better generalization while maintaining higher model capacity. Extensive experiments on several benchmarks highlight MRLs' ability to perform at par or better than competing techniques. Also, MRL's design versatility allows for incorporating components like Group Convolutions which can be used to exploit additional dataset symmetries for better generalization. 

\bibliography{}\bibliographystyle{plain}

\appendix

\section{Appendix}

\subsection{ImageNet Classification}\label{Appen:IC}

Table \ref{table:cvtmodel} describes the network architecture used for ImageNet classification experiments. 
\begin{table}[htbp] 
  \centering
  \caption{$L$ denotes the number of blocks, $D$ denotes the length of the hidden dimension maintained throughout the block with the hidden dimension per head fixed to 64, and $r$ denotes the window size of the regions of the MRL block. We use a depthwise convolution with one filter per channel for the Mixing Locally part. The expansion rate for the inverted bottleneck is always 4.} \label{table:cvtmodel}
    \begin{tabular}{c|c|c|c}
    \toprule
    Stages & Size  & MRL-CvT-13 & MRL-CvT-21 \\
    \midrule
    S0    & 1/4   & $L_0 = 1, D_0 = 64, r_0 = 4 $ & $L_0 = 1, D_0 = 64, r_0 = 4$  \\
    \midrule
    S1    & 1/8   & $L_1 = 2, D_1 = 192, r_1 = 4$ & $L_1 = 4, D_1 = 192, r_1 = 4$ \\
    \midrule
    S2    & 1/16  & $L_2 = 10, D_2 = 384, r_2 = 2$ & $L_2 = 16, D_2 = 384, r_2 = 2$ \\
    \end{tabular}%
\end{table}%

\subsection{Practical MRL}\label{subsec:implement}
While Section \ref{subsec:MixRegion} enables an efficient feature mixture of global scale features, however, in Step \ref{step:3}, for performing self-attention, we observe parameter redundancy and input-dependent high computational cost due to linear layers for calculating the QKV value set. To allow for a more parameter efficient and computationally efficient representation of the QKV value set, we propose \textit{CommonQKV}, as explained in the following.

\paragraph{CommonQKV.} \label{CQ} For generating QKV values for the (multi-head)self-attention (MHA) layer, generally, linear transformation (linear regression) is performed corresponding to each QKV pair for each head in the MHA layer. These linear layers account for $O((hd_{h})^{2})$ parameter requirement, where $h$ is the number of heads in the MHA layer and $d_{h}$ is the embedding vector length of each head input token. These linear layers also contribute to $O(n^2(hd_h)^3)$ time complexity, which increases quadratically with the input spatial dimension, $n$, and cubic in number of total channels, $hd_h$. The effect of these parameters on computing and memory requirements becomes pronounced for networks with greater depth and large input sizes to the network. The core idea of this proposed method is to have QKV values share a common basis set (a function of the Transformer block input), using which calculating QKV is cheaper (fewer parameters) and efficient (low computational and memory cost). We transform the channel dimension of the input feature map for generating the QKV value set, using a linear transformation to create the basis set. Following the generation of a basis set, by processing the basis set by a depthwise convolution layer, individually for Q, K, and V, we generate the QKV value set for the given input feature set. By using \textit{CommonQKV}, we can reduce the time complexity and parameter requirement for computing the QKV value set by one-third, as the computational complexity due to the linear layer is limited to generating the basis set only, with limited overhead due to depthwise convolutions. 



\begin{table}[htbp]
  \centering
  \caption{Model performance (Top-1 classification accuracy) on ImageNet. $1K$ denotes training on ImageNet-1K; {\fontfamily{cmss}\selectfont $1K_{384}$} denotes pre-training on ImageNet-1K followed by finetuning on ImageNet-1K with an input size of 384; {\fontfamily{cmss}\selectfont SA/MRL} denotes the Parameter (in Millions) and FLOPs (in Giga-FLOPs) corresponding to SA or MRL blocks of the network; {\fontfamily{cmss}\selectfont  images/s} correspond to \textbf{training throughput} on one DGX-A100 Node, with the subscript denoting the mini-batch size used per GPU (provided to highlight cases where a large-batch size could could not be processed for CvT models due to out-of-memory errors); {\fontfamily{cmss}\selectfont CQ+MRL} denotes models using the proposed CommonQKV \ref{CQ} and MRL, together.} \label{table:ImagentresultsAppendix}
    \begin{tabular}{c|l|cc|cc|c|c}
    \toprule
    \multicolumn{1}{c}{} &       & \multicolumn{2}{c|}{Parameters (M)} & \multicolumn{2}{c|}{FLOPs (G)} & \multicolumn{1}{c}{Top-1} & \\
    \multicolumn{1}{c|}{Dataset} & \multicolumn{1}{c|}{Model} & SA/MRL & Full  & SA/MRL & Full  & \multicolumn{1}{c|}{Accuracy} & images/s \\
    \midrule
    \multirow{6}[4]{*}{$1K$} & \multicolumn{1}{c|}{CvT-13} & 6.40  & 19.98 & 1.42  & 4.50  & 81.6  & 496.4 \\
          & \multicolumn{1}{c|}{MRL-13} & 6.40  & 19.98 & 0.63  & 3.71  & \textbf{81.63} & 555.9 \\
          & \multicolumn{1}{c|}{CQ+MRL-13} & \textbf{3.35} & 16.93 & \textbf{0.46} & 3.54  & 81.54 & \textbf{600.2} \\
\cmidrule{2-8}          & \multicolumn{1}{c|}{CvT-21} & 10.34 & 31.54 & 2.17  & 7.13  & 82.3  & 308.6 \\
          & \multicolumn{1}{c|}{MRL-21} & 10.35 & 31.55 & 1.03  & 5.98  & \textbf{82.3} & 346.3 \\
          & \multicolumn{1}{c|}{CQ+MRL-21} & \textbf{5.42} & 26.62 & \textbf{0.78} & 5.74  & 82.1  & \textbf{365.7} \\
    \midrule
    \multirow{6}[3]{*}{$1K_{384}$} & \multicolumn{1}{c|}{CvT-13} & 6.40  & 19.98 & 7.10  & 16.30 & 83    & 132.2 \\
          & \multicolumn{1}{c|}{MRL-13} & 6.40  & 19.98 & 1.99 & 11.19 & \textbf{83.1} & 172.4 \\
          & \multicolumn{1}{c|}{CQ+MRL-13} & 3.35  & 16.93 & \textbf{1.39}  & 10.59 & 81.89 & \textbf{185.4} \\
\cmidrule{2-8}          & \multicolumn{1}{c|}{CvT-21} & 10.34 & 31.54 & 10.20 & 24.90 & 83.3  & $81.7_{32}$ \\
          & \multicolumn{1}{c|}{MRL-21} & 10.35 & 31.55 & 3.23  & 17.93 & \textbf{83.3} & $107.8_{64}$ \\
          & CQ+MRL-21 & 5.42  & 26.62 & \textbf{2.52} & 17.22 & 83.18 & \boldmath{}\textbf{$115.4_{64}$}\unboldmath{} \\
    \end{tabular}%
\end{table}%

\begin{table}[htbp]
  \centering
  \caption{Model performance on ImageNet-1K, training with input size of $224 \times 224$, utilizing CommonQKV for generating QKV value set for CvT models.}\label{table:CQCvT}
    \begin{tabular}{c|c}
    \toprule
    Model & Accuracy \\
    \midrule
    CQ+CvT-13 & 81.54 \\
    CQ+CvT-21 & 82.1 \\
    \end{tabular}%
\end{table}%

Table \ref{table:ImagentresultsAppendix} further provides comparison results of utilizing the proposed CommonQKV \ref{CQ} together with MRL for computing the QKV value set. Using CommonQKV brings in additional FLOP and training throughput improvement, up to 68\% and 20\% respectively for the pre-training phase and 75\% and 42\% for the fine-tuning phase with a larger input size. However, using CommomQKV leads to a drop of up to 0.2\% accuracy during the pretraining phase, highlighting the dis-advantage brought in by parameter-shallow representation, also studied by \cite{DBLP:journals/corr/CvT}, by reducing the feature set of K and V value set to reduce computation. Additionally, to establish the reason for the accuracy drop, we do an ablation study by utilizing CommonQKV to create the QKV value set for performing Full SA in CvT models, presented in Table \ref{table:CQCvT}. Comparing results in Table \ref{table:Imagentresults} and Table \ref{table:CQCvT}, we can see a similar drop in accuracy corresponding to models utilizing CommonQKV. This drop in the accuracy suggests that MRL, for these cases, perfectly models the role played by SA, and the accuracy drop noticed with CommonQKV is due to the under-representation of the QKV value set. 

\subsection{Experiments}
\subsubsection{HoVerNet} \label{Appen:HoverNet}

\begin{figure}[ht]
    \centering
    \includegraphics[width=12cm]{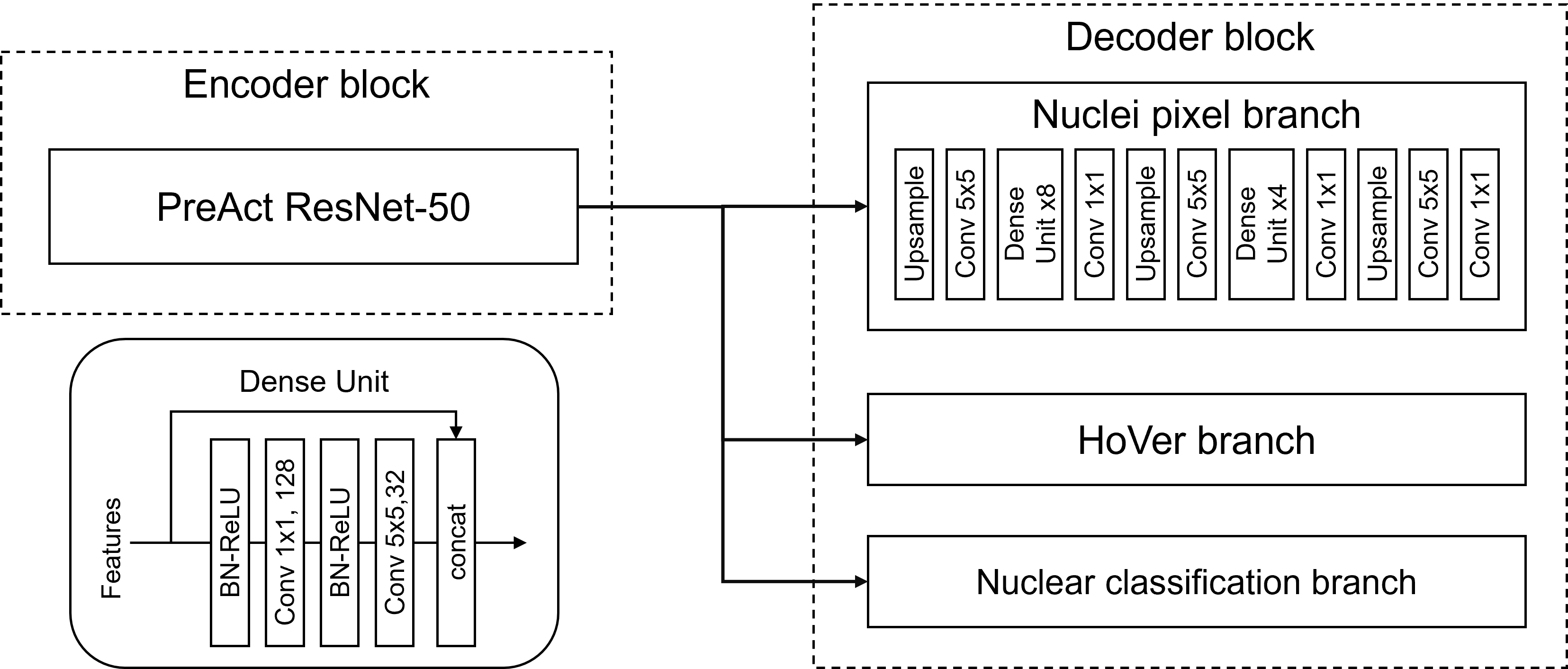}
    \caption{Original HoVerNet architecture}\label{fig:hovernet_arch}
\end{figure}

HoVerNet is one of the state-of-the-art CNN models that perform nuclei instance segmentation.HoVerNet's architecture comprises Encoder and Decoder, typical for image segmentation tasks.
The original architecture of HoVerNet adopts ResNet-50-based architecture as an encoder whose purpose is extracting features from input images.
The decoder aims to detect nuclei regions while upsampling input features. the decoder network is a group of decoder networks branches, namely, (i) nuclear pixel (NP) branch (upper one in Figure \ref{fig:hovernet_arch}); (ii) HoVer branch (middle one in figure \ref{fig:hovernet_arch}), and (iii) nuclear classification (NC) (lower one in figure \ref{fig:hovernet_arch}). The NP branch predicts whether a given pixel corresponds to nuclei or the background, the HoVer branch predicts the horizontal and vertical distances from the nuclei's center of mass to detect and separate overlapping nuclei instances, and the NC branch assigns class labels of nuclei for reach pixel. For the current task, only the NP branch and the HoVer branch are utilized as they jointly achieve instance segmentation.

\begin{figure}[ht]
    \centering
    \includegraphics[width=12cm]{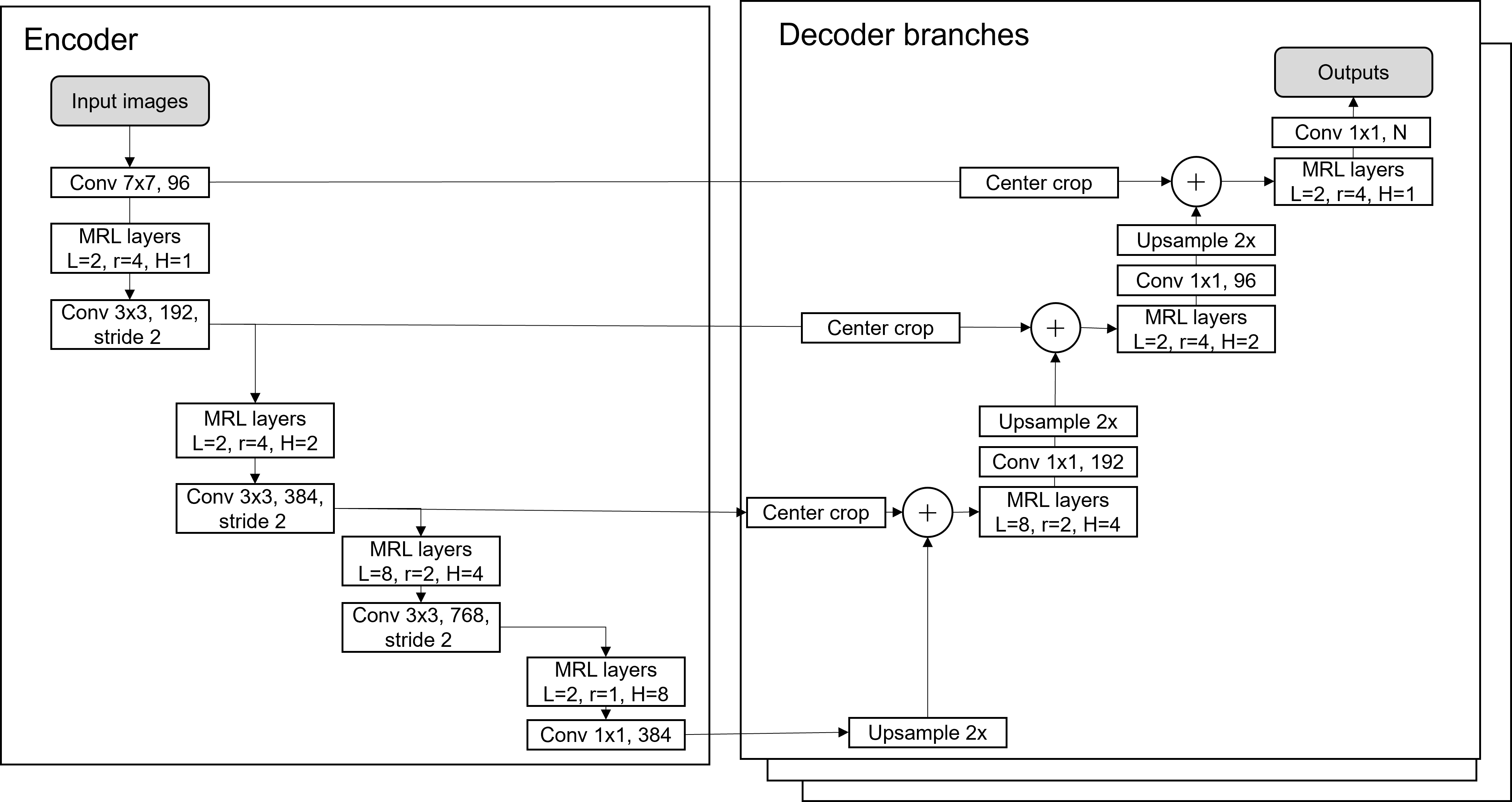}
    \caption{MRL-HoVerNet architecture. \textit{L} corresponds to number of layers; \textit{r} corresponds to the region size; \textit{H} corresponds to the number of heads, where the hidden dimension is fixed to 96.}\label{fig:MRL_hovernet_arch}
\end{figure}

In this paper, we utilize only the basic ideas of the original HoVerNet architecture that exploits the encoder's output for the several decoder branches to enhance inference quality.
Thus we replaced the whole internal architecture with our MRL-based Transformers, as shown in Figure \ref{fig:MRL_hovernet_arch}. We call this architecture "MRL-HoVerNet."

\subsubsection{Datasets} \label{appendix:datasets}

We utilized the following publicly available histopathology image datasets, CoNSeP\cite{DBLP:journals/corr/hovernet}, 
CPM17\cite{CPM17}, Kumar\cite{kumar}, and Lizard\cite{graham2021lizard}. We used Lizard only for pretraining HoVerNet based networks and we trained and evaluated the networks with the other four datasets respectively.

\begin{description}
    \item[CoNSeP] The CoNSeP dataset consists of 41 H\&E (Hematoxylin \& Eosin) stained images, each of size 1,000×1,000 pixels at 40× magnification, containing 24,319 exhaustively annotated nuclear regions with associated class labels.
    The images were extracted from 16 colorectal adenocarcinoma (CRA) WSIs, each belonging to an individual patient, and scanned in the department of pathology at University Hospitals Coventry and Warwickshire, UK.
    \item[CPM17] The CPM17 dataset consists of 64 H\&E stained images, each of size 500x500 to 600x600 pixels at 20x or 40x magnifications, containing 7570 annotated nuclear regions. The images were extracted from four different organs' WSIs of TCGA database.
    \item[Kumar] The Kumar dataset contains 30  H\&E stained images, each of size 1000x1000 pixels at 40x magnification. Within each image, the boundary of each nuclear region is fully annotated. The images were extracted from seven organs' WSIs (6 breast, 6 liver, 6 kidney, 6 prostate, 2 bladder, 2 colon and 2 stomach) of TCGA database. 
    \item[Lizard] The Lizard dataset consists of 291 H\&E stained images, each of average size 1016x917 pixels at 20x magnification, containing 495,179 annotated nuclear regions with associated class labels. The images were extracted CRA images of six other datasets. Lizard's data sources include the CoNSeP dataset, so that we carefully removed CoNSeP related images at pretraining phase.
\end{description}

\subsection{Related Works (Extended)}\label{Appen:Related_work}

Presented herewith are the some core distinctions between are presented MRL layer and some of the other works utilizing Convolutions and Self-Attention as part of their neural network design.

\begin{itemize}
    \item ConVit\cite{convit}:
    \begin{itemize}
        \item ConViT implements convolutions through the use of Gated Self-Attention.
        \item Though this allows for better capture of local dependencies however in contrast to MRL, there is no reduction in the computational requirement of the network.
        \item Additionally, MRL allows for a more dynamic design by incorporating components like Group-Convolutions.
    \end{itemize}

    \item UTNet\cite{UTNet}:
    \begin{itemize}
        \item UTNet at its base is an UNet architecture augmented with SA.
        \item The core difference between UTNet and MRL is evident in the way SA and convolution are used as part of the NN. In UTNet convolution and SA (MHSA layer) are stacked. However, in MRL, convolution, and SA are stacked as part of layers serving specific functions described in the manuscript.
    \end{itemize}
    
    \item CvT\cite{DBLP:journals/corr/CvT}:
    \begin{itemize}
        \item The core difference between CvT and MRL is very obvious by the fact that though CvT does use convolutions and SA, CvT uses convolutions for do-away with position embeddings, down-sampling, \& generating QKV values, however, uses full-SA in its MHSA layer.
        \item Instead in MRL-based CvT like NN architecture, the network can use convolutions as CvT does, while using MRL layer design as the way to mix features replacing the MHSA layer.
    \end{itemize}
    
    \item ViTAE\cite{xu2021vitae}:
    \begin{itemize}
        \item ViTAE utilizes the SA and convolutions for a similar purpose as we do, that is to capture long-scale dependencies using SA and local-scale dependencies using convolutions.
        \item However, there is a subtle difference between MRL and VITAE in how those captured dependencies are processed which changes the design complexity of the two. In ViTAE local and long-scale dependencies are processed separately without any information exchange between the two, requiring the various levels of dilation as described in their Pyramid reduction module to capture dependencies interaction.
        \item However, in the case of MRL, the local scale dependencies are captured/mixed informed by captured/mixed long-scale (Step 4 of Section 2.2). This reduced the design complexity of MRL in comparison with ViTAE.
    \end{itemize}
    
    \item CeiT\cite{ceit}:
    \begin{itemize}
        \item The difference between CeiT and MRL is in where the convolutions are used.
        \item In CeiT, convolutions are used in the Feed-Forward network of the encoder block leaving the MHSA module unchanged, whereas, in MRL-based networks, MHSA is replaced by the MRL layer while leaving the feed-forward network unchanged.
        \item This changes the way local-long scale dependencies are captured and information is exchanged. MRL is more computationally efficient as MRL reduces the computational costs associated with MHSA, while CeiT brings in additional computational costs through their LeFF layer.
    \end{itemize}
    
    \item CoAtNet\cite{DBLP:journals/corr/CoAtNet}:
    \begin{itemize}
        \item The core difference between Coatnet and MRL is in how a hybrid of SA and convolution is made.
        \item In Coatnet, uses a 'pre-normalization relative attention' wherein Attention weights are jointly decided by input-adaptive component and translation equivariant convolutional kernel. This serves the same purpose as the MRL block however at an added cost of doing convolution for all the input features.
        \item However, in the case of MRL, we do away with that additional cost by exploiting the fact that regional relations can be captured cheaply by doing down-sampling, reducing the cost of a typical MHSA layer.
    \end{itemize}
    
    \item MUSE\cite{muse}:
    \begin{itemize}
        \item Other than the fact MUSE is for NLP tasks, the core difference is how local and global information is used.
        \item In MUSE global information does not influence local information, however, in MRL local mixing is influenced by the information mixed at the regional scale. This helps to add additional context to the local feature mix.
    \end{itemize}
    
\end{itemize}

\end{document}